%% file: main.tex
\pdfoutput=1

\documentclass[11pt]{article}

\usepackage[]{EMNLP2023}

\usepackage{times}
\usepackage{latexsym}

\usepackage[T1]{fontenc}

\usepackage[utf8]{inputenc}

\usepackage{microtype}

\usepackage{inconsolata}

\usepackage{graphicx}
\usepackage{amsmath}
\usepackage{amssymb}
\usepackage{wasysym}
\usepackage{booktabs}
\usepackage{multirow}
\usepackage{colortbl}
\usepackage{makecell}
\usepackage{soul}
\usepackage{xspace}
\usepackage{siunitx}
\usepackage[linesnumbered,ruled,vlined]{algorithm2e}
\usepackage{enumitem}

\definecolor{memoryorange}{RGB}{236,128,61}

\definecolor{allgreen}{rgb}{0.6, 0.73, 0.45}
\definecolor{se3blue}{rgb}{0.61, 0.77, 0.89}
\definecolor{boldcolor}{rgb}{0,0,0}

\definecolor{sampleboldcolor}{rgb}{1.0,0.13,0.32}

\newcommand{\hlc}[2][yellow]{{%
    \colorlet{foo}{#1}%
    \sethlcolor{foo}\hl{#2}}%
}

\newcommand{\ensuretext}[1]{#1}

\newif\ifcomments
\ifcomments
\newcommand{\draftcomment}[3]{\ensuretext{\textcolor{#2}{[\ensuretext{\textcolor{#2}{\ensuremath{\textsc{#1}}}} #3]}}}
\else
\newcommand{\draftcomment}[3]{}
\fi

\newcommand{\model}{\textsc{AWESOME}\xspace}

\title{\model: GPU Memory-constrained Long Document Summarization \\ using Memory Mechanism and Global Salient Content}

\author{Shuyang Cao \and Lu Wang \\
  Computer Science and Engineering \\
  University of Michigan \\
  Ann Arbor, MI \\
  \texttt{\{caoshuy, wangluxy\}@umich.edu}}

\begin{document}
\maketitle

\input{00_abstract.tex}
\input{01_introduction.tex}
\input{02_approach_taxonomy.tex}
\input{04_proposed_method.tex}

\input{05_experiment_setup.tex}
\input{06_results.tex}
\input{08_conclusion.tex}

\input{100_limitations.tex}

\input{101_ethics_statement.tex}

\bibliography{custom}
\bibliographystyle{acl_natbib}

\appendix

\input{appendix_se3}
\input{appendixA_additional_result.tex}

\input{appendixB_dataset.tex}
\input{appendixC_implementation.tex}

\end{document}

%% file: 00_abstract.tex
\begin{abstract}

Long document summarization systems are critical for domains with lengthy and jargon-laden text, yet they present significant challenges to researchers and developers with limited computing resources. 
Existing solutions mainly focus on efficient attentions or divide-and-conquer strategies. The former reduces theoretical time complexity, but is still memory-heavy. The latter methods sacrifice global context, leading to uninformative and incoherent summaries. 
This work aims to leverage the memory-efficient nature of divide-and-conquer methods while preserving global context. 
Concretely, our framework \model uses two novel mechanisms: 
(1) \textit{External memory mechanisms} track previously encoded document segments and their corresponding summaries, to enhance global document understanding and summary coherence. 
(2) \textit{Global salient content} is further identified beforehand to augment each document segment to support its summarization. 
Extensive experiments on diverse genres of text, including government reports, meeting transcripts, screenplays, scientific papers, and novels, show that \model produces summaries with improved informativeness, faithfulness, and coherence than competitive baselines on longer documents,
while having a smaller GPU memory footprint.
    
\end{abstract}

%% file: 01_introduction.tex
\section{Introduction}

Large pre-trained transformer models have demonstrated impressive performance across popular abstractive summarization benchmarks~\cite{lewis-etal-2020-bart,2020t5}. 
Yet, transformer's quadratic \textbf{memory complexity} presents challenges for summarizing long documents with more than hundreds of words, such as scientific papers and investigation reports~\cite{cohan-etal-2018-discourse, huang-etal-2021-efficient}, making it infeasible for researchers and developers with limited hardware resources (e.g., GPUs with insufficient memories) to contribute to this important research field. 

The NLP community has made several innovations to address the long document challenge. Prior work divides a document into smaller chunks and summarizes each separately~\cite{gidiotis2020divide}, reduces the complexity of attention calculations~\cite{beltagy2020longformer}, and removes unimportant content before running an abstractor~\cite{pilault-etal-2020-extractive}.
In terms of memory efficiency, divide-and-conquer methods obtain the most significant advantage~\cite{moro2022se3}. However, information outside of a document segment and their corresponding summaries become inaccessible, leading to uninformative and incoherent summaries. 
Unsurprisingly, state-of-the-art performance is obtained by models that can maintain global context, e.g., by combining global attentions with local attentions in transformer-based summarization models~\cite{zaheer2021big, phang2022investigating}. Yet, they still require a large GPU memory footprint in practice.\footnote{These approaches require a GPU memory of >40GB to process documents with over 8K tokens, while the most cost-effective GPUs only have 24GB of memory~\cite{li_2022}.} 
Though large language models like GPT-4~\cite{openai2023gpt4} are trained to handle up to 32K tokens, the privacy and security of data transmitted and shared through the API remain concerning, particularly in sectors dealing with sensitive information, e.g., clinical notes. Local model development can bolster privacy and security; however, limited computational resources in these scenarios necessitate the exploration of efficient modeling techniques. 

Therefore, this work aims to address the problem of long document summarization using constrained resources, specifically focusing on \textit{constrained GPU memory}. 
We propose \textbf{\model}\footnote{Our code will be made available at \url{https://shuyangcao.github.io/projects/awesome/}}, which is built on the memory-efficient divide-and-conquer approach, and \underline{A}ugmented \underline{W}ith \underline{E}stimated \underline{S}alient c\underline{O}ntent and \underline{ME}mory mechanism. In essence, \model maintains global context of both the source document and the summary generated so far with a limited memory usage, to enhance summary informativeness, faithfulness, and coherence.

First, \textbf{external memory mechanism} is used on the encoder side of \model to store information as it reads in document segments in sequence. This maintains relevant context for improved document understanding and salient content detection, thus promoting summary informativeness and faithfulness. 
Another memory is applied on the decoder side to improve generation coherence by tracking the partial summaries generated for prior document segments. 
Importantly, to ensure the GPU memory efficiency of \model, we \textit{curb gradients from propagating to other document and summary segments} and only allow a limited number of layers to maintain the external memory.

Second, \model incorporates \textbf{global salient content} selected by an efficiently trained extractor through (1) direct text concatenation, or (2) inserting their key-value matrices into attention calculation. 
This lets the summarizer be aware of important topics at a global level, to enhance salience estimation and summary informativeness. 

We experiment with five popular long-input benchmarks of different genres: investigation reports in GovReport~\cite{huang-etal-2021-efficient}, meeting transcripts in QMSum~\cite{zhong-etal-2021-qmsum}, TV screenplays in SummScreen~\cite{chen-etal-2022-summscreen}, scientific papers in arXiv~\cite{cohan-etal-2018-discourse}, and fictions in BookSum~\cite{kryscinski-etal-2022-booksum}. 
First, on all the five datasets, all \model variants uniformly outperform Se3~\cite{moro2022se3}, the divide-and-conquer baseline, on summary informativeness as evaluated by ROUGE~\cite{lin-2004-rouge} and on coherence as measured by DiscoScore~\cite{zhao2022discoscore} and a metric based on entity graphs~\cite{guinaudeau-strube-2013-graph}---both metrics are highly correlated with human judgment, according to~\newcite{zhao2022discoscore}. 
Second, \model with memory mechanisms also improves summary faithfulness over Se3 on GovReport, according to SummaC~\cite{laban-etal-2022-summac}, an entailment-based faithfulness metric. 
Lastly, compared with more memory-intensive models that also maintain global context, such as \newcite{phang2022investigating} and \newcite{liu2022page}, \model achieves higher automatic scores for informativeness, coherence, and faithfulness on GovReport~\cite{huang-etal-2021-efficient}.
On BookSum which comprises the lengthiest documents and summaries among the five datasets, \model produces more informative and coherence outputs than recent models. 

%% file: 02_approach_taxonomy.tex
\section{Related Work}
\label{sec:related_work}

\subsection{Efficient Long Document Summarization}
\label{sec:related_work_longdoc}

\begin{table}[t]
    \centering
    \small
    \setlength{\tabcolsep}{4pt}
    \begin{tabular}{lcccc}
    \toprule
        \textbf{Approach} & \textbf{In$\rightarrow$Out} & \textbf{Enc} & \textbf{Enc$\leftarrow$Dec} & \textbf{Dec} \\
        \midrule
        Efficient Attention & $x \rightarrow y$ & $\blacksquare$ & $\blacksquare$ & $\CIRCLE$ \\
        \midrule
        Extract-Abstract & $x_e \rightarrow y$ & $\square$ & $\bigstar$ & $\CIRCLE$ \\
        \midrule
        Dynamic Weight & $x \rightarrow y$ & $\square$ & $\blacksquare + \bigstar$ & $\CIRCLE$ \\
        \midrule
        Divide-Conquer & $x_i \rightarrow y_i$ & $\square$ & $\square$ & $\Circle$ \\
        \bottomrule
    \end{tabular}
    \caption{
        Existing approaches to long document summarization (\S\ref{sec:related_work_longdoc}). 
        \textbf{In$\rightarrow$Out}: Longer inputs ($|x| > |x_e| > |x_i|$) or outputs ($|y| > |y_i|$) produce more nodes in the computation graph, thus the higher memory consumption. 
        \textbf{Enc}: Encoder accessing partial documents ($\square$) hurts document understanding, compared to reading the full text ($\blacksquare$). 
        \textbf{Enc$\leftarrow$Dec}: Decoder reading the full document ($\blacksquare$) or pre-identified salient content ($\bigstar$) enhances summary informativeness, compared to a segment ($\square$). 
        \textbf{Dec}: Decoder accessing previously generated summary content ($\CIRCLE$) is crucial for generation coherence than reading a current summary segment only ($\Circle$).
    }
    \label{tab:approach_taxonomy_concise}
\end{table}

We categorize existing efficient long document summarization models into four major types, as summarized in Table~\ref{tab:approach_taxonomy_concise}. 
The model \textbf{input} can be an original document, extracted important segments of the document, or a document segment, which are denoted as $x$, $x_e$, or $x_i$ (for the $i$-th segment), and typically, $|x| > |x_e| > |x_i|$. 
The \textbf{output} can be the full summary $y$ or a summary segment $y_i$ (for $x_i$), where $|y| > |y_i|$. 
Importantly, longer inputs and outputs expand larger computation graph, leading to higher GPU memory usage. 
Moreover, we analyze both the \textbf{document context} and the \textbf{summary context} used by each approach when generating summaries. Specifically, we check 
(1) full vs. partial documents that are consumed to obtain the encoder representations (\textbf{Enc}); 
(2) full vs. partial encoder representations that are attended by the decoder (\textbf{Enc$\leftarrow$Dec}); 
and (3) full vs. partial output that is accessed by the decoder (\textbf{Dec}). 

\textbf{Efficient attentions} are designed to reduce the quadratic complexity of the original transformer architecture~\cite{vaswani2017attention} and maintain full encoding context by combining global attentions with local attentions built on sliding windows~\cite{beltagy2020longformer, zaheer2021big}, text blocks~\cite{phang2022investigating,tay2020sparse}, or clusters of similar tokens~\cite{Kitaev2020Reformer,roy-etal-2021-efficient}. 
Besides the aforementioned attention variants designed for self-attentions, recent work has reduced the memory usage of decoder cross attentions by distributing encoder outputs to different attention heads~\cite{huang-etal-2021-efficient} or selecting attendable encoder outputs via KNN search~\cite{bertsch2023unlimiformer}.
Despite the reduced complexity, efficient attention-based systems effectively require reading the full document $x$ to generate a summary $y$ during model training and thus still need huge GPU memory that scales with the input length. 

\textbf{Extract-then-abstract} systems circumvent the long sequence challenge by first identifying the salient segments, $x_e$ (e.g., sentences), using an extractor, and then running an abstractor over $x_e$ to produce the final summary~\cite{pilault-etal-2020-extractive, liu-lapata-2019-hierarchical, zhao2020seal}. 
However, the extracted segments may contain incomplete and out-of-context information that leads to incomprehensible and unfaithful summaries.

To mitigate the error propagation issue of a two-stage approach, recent studies bridge the extractor and abstractor via \textbf{dynamic weights} over document segments. Rather than feeding the extracted segments directly to the abstractor, at each summary decoding step, DYLE~\cite{mao-etal-2022-dyle} first predicts an output token distribution for each segment separately, and then aggregates over all the extracted segments as weighted by their extraction salience. 
PageSum~\cite{liu2022page} further alleviates context loss by averaging decoder output representations conditioned on all document segments. 
Though their abstractor processes each document segment $x_i$ separately, jointly training the extractor and the abstractor still requires loading the full document $x$ into the GPU memory.

\textbf{Divide-and-conquer} systems split a document into multiple non-overlapping segments and summarize each segment separately, as done in~\newcite{gidiotis2020divide} and Se3~\cite{moro2022se3}. 
Summ$^N$~\cite{zhang-etal-2022-summn} uses an additional summarization stage to further condense the segmented summaries. 
As each document segment $x_i$ is summarized separately, the divide-and-conquer approach's fixed GPU memory footprint is independent from the document length. This fits well with our goal of long document summarization with limited memory. 
However, without access to other parts of the document and their summaries, the summarizer struggles for content salience estimation in each isolated segment,
and generates incoherent outputs when piecing together summaries. 
Though \citet{wu2021recursively} concatenate previously generated summaries as part of the input, a complicated strategy is required for training sample construction.

\model is built on the memory-efficient divide-and-conquer approach, and improves summary informativeness, coherence, and faithfulness by using newly designed external memories for accumulating salient information from other document segments and their generated summaries. We further augment \model with global salient content to provide important topics at the document level, when summarizing each segment. 

\subsection{Memory and Content Augmentation}

Different memory mechanisms have been studied for long-range text \textit{understanding} tasks. 
For instance, Transformer-XL~\cite{dai-etal-2019-transformer} caches intermediate representations produced in the last document segment and attends over these representations. 
Compressive Transformer~\cite{Rae2020Compressive} further increases the context range by compressing the oldest cached representations. 
To simulate memory reading and writing, Recurrent Memory Transformer~\cite{bulatov2022recurrent} includes extra memory vectors in each text segment and passes their corresponding output vectors to the next segment.
Instead of using a memory with a fixed size, Memorizing Transformer~\cite{wu2022memorizing} stores all prior representations as key-value pairs, and performs an approximate kNN lookup to retrieve representations to augment the current segment. 
However, existing work on memory mechanisms focuses on language modeling, while \textit{incorporating memory mechanisms into the decoding process for generation tasks is nontrivial} as it requires updating both decoding states (e.g., beams) and memory states.
Our work is the first to leverage memory mechanisms and content augmentation to incorporate global context for the purpose of memory-efficient long document summarization.

%% file: 04_proposed_method.tex
\section{External Memory and Global Salient Content Augmentation}

\begin{figure}[t]
    \centering
    \includegraphics[width=0.49\textwidth]{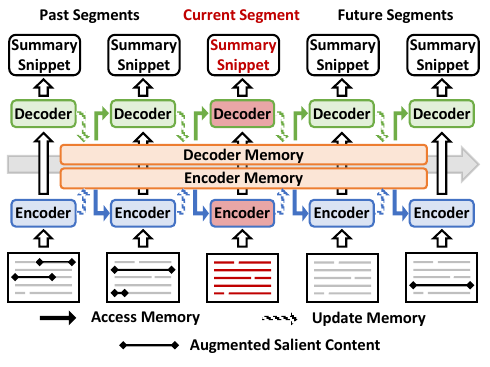}
    \caption{
    Illustration of \model. Encoder and decoder \textcolor{memoryorange}{\textbf{memories}} can be accessed any time and updated after reading each document segment and generating the corresponding summary. They accumulate global context that improves summary informativeness and coherence (\S\ref{sec:external_memory}). 
    When encoding each segment, global salient content from other segments (lines with $\blacklozenge$-shaped ends, from both past and future) are provided to further assist salience estimation (\S\ref{sec:global_content_augment}). 
    }
    \label{fig:model_illustration}
\end{figure}

The architecture of \model (Figure~\ref{fig:model_illustration}) is based on Se3~\cite{moro2022se3}, where a document is summarized segment by segment, with the final summary obtained by concatenating the resultant summaries.
Document sentences are split into segments with up to 768 tokens each, while reference summary sentences are assigned to their most overlapping segment to create the oracle summary, as detailed in Appendix~\ref{appendix:se3}.
Following Longformer~\cite{beltagy2020longformer}, we initialize the encoder and decoder parameters from BART~\cite{lewis-etal-2020-bart}.
\model preserves the global context and builds communications across segments with minimal GPU memory increase, by (1) employing external memories in \textit{both the encoder and the decoder} to gather relevant information (\S\ref{sec:external_memory}), and (2) augmenting \textit{the encoder} with salient content from other segments (\S\ref{sec:global_content_augment}).

\subsection{External Memory Mechanisms}
\label{sec:external_memory}

We design two external memory mechanisms to efficiently enable the information flow from prior segments to the current segment. 
Specifically, each memory module maintains a matrix $M \in \mathbb{R}^{m \times d}$, where $m = 1024$ is the memory size and $d = 1024$ is the hidden state dimension of BART. 
$M$ is updated after encoding each document segment and then passed to the next segment. We denote the memory matrix after the $t$-th segment as $M^t$. 
Each layer of the encoder and decoder can be equipped with one such external memory.
Below we describe two mechanisms to update $M^t$ and incorporate it in both the encoding and decoding processes. 
The layer index in the formulas is omitted for simplicity.

\smallskip\noindent
\textbf{Compressive Memory.}
For each document segment, compression-based memory caches its input vectors to be fed into self-attention calculation. 
Since storing the input vectors as-is requires the memory usage $m$ to scale linearly with the context length, we dedicate half of $M^t$ to store the compressed memory, with a compression ratio of $r$.  
With $H^t_{inp}$ denoting the matrix that contains input vectors to the transformer self-attention, the memory compression and update processes are:

{\fontsize{10}{12}\selectfont
    \begin{align}
        M^{t-1}_c, M^{t-1}_u &= M^{t-1}[:\frac{m}{2}],  M^{t-1}[\frac{m}{2}:] \\
        M'_u &= \mathrm{concat} (M^{t-1}_u, \mathrm{SG}(H^t_{inp})) \\
        M'_c &= \mathrm{compress} (M'_u [:-\frac{m}{2}]) \\
        M^t_u &= M'_u[-\frac{m}{2}:] \\
        M^t_c &= \mathrm{concat}(M^{t-1}_c, M'_c)[-\frac{m}{2}:] \\
        M^t &= \mathrm{concat} (M^t_c, M^t_u)
    \end{align}
}
where $\mathrm{SG}(\cdot)$ denotes stopping the gradient backpropagation to lower GPU usage, and $\mathrm{compress}(\cdot)$ performs convolutions with their stride and kernel size set to the compression ratio $r$. $r$ is set to $5$ after tuning on the development sets. 

Next, to leverage the memory from the previous segment in summarizing the current segment, $M^{t-1}$ is concatenated with the inputs to the self-attentions to obtain the key-value matrices: 

{\fontsize{10}{12}\selectfont
    \begin{align}
        H^t_{mem} &= \mathrm{concat}(M^{t-1}, H^t_{inp}) \\
        H^t_{self} &= \mathrm{Attn} (\underbrace{H^t_{inp}}_{query}, \underbrace{H^t_{mem}}_{key}, \underbrace{H^t_{mem}}_{value})
    \end{align}
}%
where $H^t_{self}$ is the output of the self-attention.

Our compression-based memory is adopted from Compressive Transformer~\cite{Rae2020Compressive}, a decoder-only model for language modeling. We are the first to apply it to both the encoder and the decoder of a Transformer model and on long document summarization tasks. 

Compressive memory favors recency, particularly the previous segment and its summary, potentially causing older relevant history to be lost during compression.

\smallskip\noindent
\textbf{Attentive Memory.}
To mitigate the recency bias by compressive memory, we further investigate an attention-based memory updating mechanism, to selectively include content in $M^t$. 
First, the memory is additionally accompanied by an extra cross-attention in each of the encoder and decoder layers, specialized in retrieving relevant information from $M^t$. 
Following prior study~\cite{lei-etal-2020-mart} that uses memories in video captioning, we update $M^t$ with a gate matrix $G^t$ to control the amount of content to be updated:

{\fontsize{10}{12}\selectfont
    \begin{equation}
        M^t = G^t \odot U^t + (1 - G^t) \odot M^{t-1}
    \end{equation}
}
where $\odot$ denotes the element-wise product and $U^t$ is the matrix containing vectors to update the memory. $U^t$ and $G^t$ are obtained as follows:

{\fontsize{10}{12}\selectfont
    \begin{align}
        U^t &= \tanh ( W_{u1} M^{t-1} + W_{u2} S^t ) \\
        G^t &= \sigma ( W_{g1} M^{t-1} + W_{g2} S^t ) \\
        S^t &= \mathrm{Attn} ( \underbrace{M^{t-1}}_{query}, \underbrace{\mathrm{SG}(H^t_{self})}_{key}, \underbrace{\mathrm{SG}(H^t_{self})}_{value} )
    \end{align}
}
where $W_{*}$ are learnable matrices, $S^t$ synthesizes the current segment via an attention calculation, and $\mathrm{SG}(\cdot)$ indicates stopping the gradient backpropagation. 
In each encoder and decoder layer, an extra cross-attention is inserted after the self-attention, where $M^{t-1}$ is attended and incorporated into the current segment's summarization process.

Unlike our approach, the memory in \citet{lei-etal-2020-mart} does not employ gradient stopping. This omission eliminates the memory efficiency gained from the divide-and-conquer strategy, leading to comparable high memory usage as the efficient attention strategy.\footnote{Without gradient stopping, the model fails to complete training with 48GB GPU memory.}
While their memory is suitable for generating short image captions, \textit{our design with gradient stopping is crucial for efficient long document summarization}.

\smallskip\noindent
\textbf{Selective Addition of External Memory.}
External memory incurs overhead in GPU memory usage. To mitigate this overhead, we consider selectively adding external memory to specific layers, as the importance of external memory varies according to the different functions of layers in the model.
Our pilot study suggests that the last layers of the Transformer model more effectively utilize external memory compared to the first layers. To avoid an exhaustive search for the optimal layer or combination of layers for each dataset, we choose to uniformly equip the last three layers with external memory across all datasets unless otherwise specified.\footnote{Compared to adding external memory to all layers, selective addition reduces GPU memory usage by about 9GB.}

\subsection{Global Salient Content Augmentation}
\label{sec:global_content_augment}

The memory mechanisms only grant access to prior content in the documents, yet subsequent context can also help with salience estimation, e.g., elaborating the pros and cons of a proposed solution makes it necessary to introduce the problem and the solution. 
Moreover, memories store content implicitly, so it is unclear whether relevant information can be stored and retrieved effectively. 
Therefore, we inform the system of a document's important sentences, which are pre-identified by a separately-trained extractor.
The details of extractor training can be found in Appendix~\ref{appendix:implementation}. 
After extracting important sentences in a document, we study two methods of injecting them into the summarizer.

\smallskip\noindent
\textbf{Text Concatenation.} 
For each segment, we include the extracted sentences in the following way to prioritize long-term context. We start with the ``outermost'' extracted sentences, i.e., the earliest sentence in the past segments and the last sentence in the future segments, and repeat this process until the input has reached the maximum length accepted by the positional encoding of the model ($1024$ for BART).\footnote{Other inclusion strategies can be explored in future work.} 
To differentiate the content in the current segment from the added sentences, 
we prefix the current segment and the added sentences from before/after the current segment with ``\texttt{Current chunk:}'', ``\texttt{Previous important sentences:}'', and ``\texttt{Next important sentences:}'', respectively.
Text concatenation is easy to implement and most compatible with the source modality, 
but the memory usage increase is quadratic to the length of the augmented content. 

\smallskip\noindent
\textbf{Key-value Vectors.} 
To circumvent the quadratic memory increase, we join the key-value representations of tokens in important sentences in the encoder self-attentions, and directly inject them into the summarizer encoder. The memory increase is only linear to the augmented content's length.

\begin{table*}[t]
    \centering
    \small
    \setlength{\tabcolsep}{4pt}
    \begin{tabular}{lllllllll}
    \toprule
        \textbf{Model} & \textbf{R-1} $\uparrow$ & \textbf{R-2} $\uparrow$ & \textbf{R-L} $\uparrow$ & \textbf{Ent Prec} $\uparrow$ & \textbf{SummaC} $\uparrow$ & \textbf{Disco} $\downarrow$ & \textbf{Ent Graph} $\uparrow$ & \textbf{GPU Mem} $\downarrow$ \\
        \midrule
        Se3 & 46.56 & 23.22 & 44.36 & 98.24 & 14.71 & 7.37 & 1.41 & 11.1 \\
        BlockAttn & 57.46 & 26.78 & 54.82 & 97.45 & 20.43 & 5.91 & \underline{2.05} & 25.6 \\
        Longformer & 57.40 & 26.92 & 54.70 & 97.52 & 20.39 & 5.68 & \underline{2.05} & 25.3 \\
        LongT5 & 54.21 & 24.87 & 51.06 & 96.41 & 13.34 & 4.81 & 1.56 & 25.4 \\
        Unlimiformer & 56.35 & 25.94 & 53.83 & 92.19 & 6.05 & 5.36 & 1.96 & 27.0 \\
        Extract-Abstract & 56.89 & 24.76 & 54.26 & 92.82 & \textbf{{\textcolor{boldcolor}{22.07}}} & 4.03 & \textbf{{\textcolor{boldcolor}{2.09}}} & 13.2 \\
        PageSum & 56.80 & 23.26 & 54.11 & 89.56 & 6.82 & \textbf{{\textcolor{boldcolor}{3.04}}} & 1.88 & 24.9 \\
        \midrule
        \multicolumn{9}{l}{\textit{\model using External Memory Only}} \\
        \quad \quad Compressive & \hlc[se3blue]{50.71}$^\dag$ & \hlc[se3blue]{23.91} & \hlc[se3blue]{48.45}$^\dag$ & 89.17 & \hlc[se3blue]{15.34} & \hlc[se3blue]{5.16}$^\dag$ & \hlc[se3blue]{1.94}$^\dag$ & 12.5 \\
        \quad \quad Attentive (Attn) & \hlc[allgreen]{{\underline{58.44}}}$^\ast$ & \hlc[allgreen]{{\underline{27.71}}}$^\ast$ & \hlc[allgreen]{{\underline{55.98}}}$^\ast$ & \hlc[allgreen]{{\textcolor{boldcolor}{\textbf{98.33}}}} & \hlc[se3blue]{18.98}$^\dag$ & \hlc[se3blue]{{\underline{3.62}}}$^\dag$ & \hlc[se3blue]{1.98}$^\dag$ & 14.0 \\
        \multicolumn{9}{l}{\textit{\model using Global Salient Content Only}} \\
        \quad \quad Text-concat (Txt) & \hlc[se3blue]{56.65}$^\dag$ & \hlc[allgreen]{27.68}$^\ast$ & \hlc[se3blue]{54.11}$^\dag$ & 97.93 & 12.23 & \hlc[se3blue]{5.05}$^\dag$ & \hlc[se3blue]{\textbf{{\textcolor{boldcolor}{2.09}}}}$^\dag$ & 12.0 \\
        \quad \quad Key-value Vectors & \hlc[se3blue]{55.02}$^\dag$ & \hlc[se3blue]{26.39}$^\dag$ & \hlc[se3blue]{52.41}$^\dag$ & 98.22 & 11.52 & \hlc[se3blue]{4.75}$^\dag$ & \hlc[se3blue]{1.75}$^\dag$ & 14.3 \\
        \model (Attn + Txt)  & \hlc[allgreen]{{\textcolor{boldcolor}{\textbf{58.76}}}}$^\ast$ & \hlc[allgreen]{{\textcolor{boldcolor}{\textbf{28.18}}}}$^\ast$ & \hlc[allgreen]{{\textcolor{boldcolor}{\textbf{56.05}}}}$^\ast$ & \hlc[se3blue]{{\underline{98.31}}} & \hlc[se3blue]{{19.22}}$^\dag$ & \hlc[se3blue]{3.86}$^\dag$ & \hlc[se3blue]{2.03}$^\dag$ & 14.8 \\
        \bottomrule
    \end{tabular}
    \caption{
    Results on GovReport. The best and second best results per metric are {\textcolor{boldcolor}{\textbf{bolded}}} and \underline{underlined}. 
    Results by \model variants that are better than all comparisons and Se3 are shaded with \hlc[allgreen]{green} and \hlc[se3blue]{blue}, respectively. 
    \model with attentive memory only and its full version that additionally uses salient content through text concatenation obtain the highest ROUGE scores (in green) and are comparable or better on faithfulness (Ent Prec \& SummaC) and coherence (Disco \& Ent Graph) than base model Se3. 
    $\ast$: our model is better than all comparisons with approximation randomization test ($p < 0.0005$); 
    $\dag$: our model is better than Se3 ($p < 0.0005$).
    }
    \label{tab:govreport_result}
\end{table*}

Concretely, the summarizer encoder first encodes all document segments and obtains the representations (i.e., encoder outputs) of tokens belonging to the extracted important sentences. 
During training, the token representations of these sentences are concatenated with the key-value matrices in the encoder self-attentions while the query matrix remains in its original form. 
Up to $1024$ tokens are concatenated via the same inclusion method for text concatenation, to prioritize the outermost sentences. 
A similar idea has been used by Memorizing Transformer~\cite{wu2022memorizing} to include retrieved text representations from past segments for long-form language modeling.
Our method differs in two aspects. 
First, we extract representations from \textit{future segments}, which are crucial for accurately identifying salient content.
Second, we apply \textit{a learnable projection} to the augmented representations prior to key-value concatenation. This process is crucial in improving compatibility with the original key-value matrices.

%% file: 05_experiment_setup.tex
\section{Experimental Setups}

\noindent
\textbf{Datasets.}
We conduct experiments on GovReport~\cite{huang-etal-2021-efficient}, QMSum~\cite{zhong-etal-2021-qmsum}, SummScreen~\cite{chen-etal-2022-summscreen}, arXiv~\cite{cohan-etal-2018-discourse}, and BookSum~\cite{kryscinski-etal-2022-booksum}. 
The average input lengths of these datasets range from 6K to 143K (Appendix~\ref{appendix:dataset_stat}).

\smallskip\noindent
\textbf{Experiment Setups and Comparisons.}
Our main experiments are conducted with a \textit{GPU memory constraint of 27GB}.
For each model, we truncate the input such that its maximum GPU memory usage during training does not exceed the constraint when gradient checkpointing~\cite{chen2016training} is \textit{disabled}.
The constraint is specifically chosen such that the baselines perform reasonably. 
Appendix~\ref{appendix:model_input} provides information on the maximum number of input tokens that can conform to the constraint for other models.

For baselines, in addition to the divide-and-conquer \textbf{Se3} model~\cite{moro2022se3}, we compare with state-of-the-art or popular long document summarization systems including \textbf{BlockAttn}~\cite{phang2022investigating}, \textbf{Longformer}~\cite{beltagy2020longformer}, \textbf{LongT5}~\cite{guo-etal-2022-longt5}, and \textbf{Unlimiformer}~\cite{bertsch2023unlimiformer}.
We also include an extract-then-abstract model (\textbf{Extract-Abstract}) and \textbf{PageSum}~\cite{liu2022page} that leverages dynamic weights, as discussed in \S\ref{sec:related_work}.
All models are initialized from \texttt{BART-large}, except for LongT5 that is pre-trained on long-form data.
Details of baseline models are reported in Appendix~\ref{appendix:implementation}.

\smallskip\noindent
\textbf{Evaluation Metrics.}
We evaluate summary \textit{\textbf{informativeness}} using ROUGE~\cite{lin-2004-rouge}. 
To measure \textit{\textbf{coherence}},
we use DiscoScore~\cite{zhao2022discoscore} (\textbf{Disco}), a reference-based metric that evaluates discourse coherence by comparing focus (e.g., nouns) frequency and semantics between the system summary and the reference.
We also report a graph-based reference-free coherence metric~\cite{guinaudeau-strube-2013-graph} (\textbf{Ent Graph}),
which measures the connectivity of summary sentences linked by entities, reflecting the coherence of topic transitions. 
For summary \textit{\textbf{faithfulness}},
we follow prior work on text generation~\cite{iv-etal-2022-fruit} and show the precision of the entities (\textbf{Ent Prec}) in the summary with respect to the document. 
Additionally, a recent model-based faithfulness metric, \textbf{SummaC}~\cite{laban-etal-2022-summac}, is used.

Finally, we show the maximum size of allocated \textbf{GPU memory} by each model during training. 

%% file: 06_results.tex
\section{Results}


We report results by all \model variants and comparison models on \textbf{GovReport} in Table~\ref{tab:govreport_result}. 
Compared with Se3, \model variants \textit{consistently achieve better performance} on both \textit{ROUGE} and \textit{coherence} scores, indicating the importance of maintaining global context for accurate salience estimation of local content and enforcing coherent transitions across segment-level summaries. This can also be demonstrated by the sample outputs in Table~\ref{tab:output sample}. 
Summaries generated by Se3 tend to be shorter, as Se3 fails to plan at a global level. 
On faithfulness, \model with attentive memory has the best entity precision among all models and also improves SummaC over Se3, while \textit{only} augmenting \model with global salient content hurts faithfulness. 
Inspecting the model outputs, we find that using attentive memory improves understanding concepts of long-term dependencies, e.g., connecting a strategy with its related information that appears earlier in the report. 

\begin{table}[t]
    \centering
    \small
    \begin{tabular}{p{0.45\textwidth}}
    \toprule
        \textbf{Se3:} VA is required to publish information on appointment wait times at each VA medical facility for primary care, specialty care, and hospital care and medical services, which it does through two public websites. VA has taken a number of actions to address deficiencies GAO found in wait-time measurement and implementation of its scheduling policy. For wait-time measurement, these actions included changes to the wait-time measurement definitions, provision and documentation of scheduler training, and improved oversight through audits, all of which have been in a state of flux for the past 6 years. On July 12, 2019, VA provided GAO additional updates on efforts to implement \textcolor{sampleboldcolor}{\textbf{GAO's related recommendations}}.\\
        \midrule
        \textbf{\model:} GAO \textcolor{sampleboldcolor}{\textbf{recommended}} that VA either clarify its scheduling policy to better define the desired date, or identify clearer wait-time measures that are not subject to interpretation and prone to scheduler error. VA concurred with the \textcolor{sampleboldcolor}{\textbf{recommendation}}, which GAO has identified as among those \textcolor{sampleboldcolor}{\textbf{recommendations}} that warrant priority attention. VA has taken a number of actions to address \textcolor{sampleboldcolor}{\textbf{GAO's recommendations}} regarding deficiencies GAO found in wait-time measurement and implementation of its scheduling policy. For wait-time measurement, these actions included changes to the wait-time measurement definitions, provision and documentation of scheduler training, and improved oversight through audits, all of which have been in a state of flux for the past 6 years.  On July 12, 2019, VA provided GAO additional updates on efforts to implement \textcolor{sampleboldcolor}{\textbf{GAO's related recommendations}}. \\
        \bottomrule
    \end{tabular}
    \caption{
    Summary snippets generated by Se3 and \model. \model's summary is more coherent, with natural transitions surrounding ``\textcolor{sampleboldcolor}{\textbf{GAO's recommendation}}'', while Se3 abruptly introduces the topic. 
    }
    \label{tab:output sample}
\end{table}

Of the two types of external memory mechanisms, \textit{attentive memory outperforms compression-based memory on all metrics}, which highlights the advantage of adaptively updating the stored context. 
Meanwhile, \textit{directly concatenating salient content with the input yields higher ROUGE scores} than injecting key-value vectors into the attention calculation, though the latter is less memory-intensive. 
We believe natural language-based augmentation better interleaves with the document segment, echoing the findings by prior work on using retrieval for question answering~\cite{wu2022efficient}.

Importantly, \textit{under a strict GPU memory constraint, \model with external memory mechanisms and global salient content augmentation achieves the best ROUGE scores} among all models, while obtaining competitive results on other measures.
Though efficient attention models and PageSum can perform remarkably when given higher-capacity GPUs as in the original work,
they generate less informative summaries when truncation is required to comply with the memory constraint, emphasizing the importance of studying memory-efficient long document summarization models.
Furthermore, with selective addition of external memory, \model adds only about 4GB of GPU memory usage, enhancing the model performance efficiently.

\begin{table}[t]
    \centering
    \small
    \setlength{\tabcolsep}{3pt}
    \begin{tabular}{llllll}
    \toprule
        \textbf{Model} & \textbf{R-1} $\uparrow$ & \textbf{R-2} $\uparrow$ & \textbf{R-L} $\uparrow$ & \textbf{Disco} $\downarrow$ & \textbf{GPU} $\downarrow$ \\
        \midrule
        Se3 & 29.28 & \underline{10.51} & 25.93 & 0.77 & 8.1 \\
        BlockAttn & 30.76 & \phantom{0}8.26 & 26.49 & 0.50 & 22.8 \\
        Longformer & 29.18 & \phantom{0}7.82 & 24.94 & 3.07 & 26.5 \\
        LongT5 & \underline{31.88} & 10.07 & \underline{27.82} & \underline{0.44} & 25.4 \\
        Unlimiformer & 30.57 & \phantom{0}8.82 & 26.89 & 0.49 & 26.9 \\
        Extract-Abstract & 17.63 & \phantom{0}5.65 & 16.02 & 4.02 & 10.3 \\
        PageSum & 29.55 & \phantom{0}7.38 & 26.11 & {\textcolor{boldcolor}{\textbf{0.31}}} & 21.5 \\
        \midrule
        \model \\
        \quad Attn Only & \hlc[allgreen]{{\textcolor{boldcolor}{\textbf{34.86}}}}$^\dag$ & \hlc[allgreen]{{\textcolor{boldcolor}{\textbf{12.69}}}} & \hlc[allgreen]{{\textcolor{boldcolor}{\textbf{31.09}}}}$^\ast$ & \hlc[se3blue]{0.68} & 12.9 \\
        \quad Attn + Txt & \hlc[se3blue]{{31.16}}$^\dag$ & 10.11 & \hlc[se3blue]{{27.66}} & \hlc[se3blue]{0.69} & 13.3 \\
        \bottomrule
    \end{tabular}
    \caption{
    Results on meeting transcripts in QMSum. Equipped with attentive memory only, \model achieves the best ROUGE scores. Though better than some baselines, adding extracted salient content does not further boost the performance, due to the low performance of the extractor on dialog data. 
    }
    \label{tab:qmsum_result}
\end{table}

On \textbf{QMSum} (Table~\ref{tab:qmsum_result}), \model with attention-based memory outperforms all comparisons on ROUGE scores. 
While our models' summaries are more coherent than the summaries by Se3, as measured by DiscoScore, the differences among all models are less pronounced compared to the ones on GovReport. This is because QMSum contains shorter summaries than GovReport (69 vs. 553), thus involving fewer topic transitions. 
We also find that the extractor performs poorly on QMSum, leading to degraded results after augmenting our model with the extracted salient content.
Specifically, the F1 score of the extractor on the test set is only $1.29$, as opposed to $27.85$ on GovReport. 
Compared to our model, the extract-then-abstract model is more prone to its errors and produce summaries of the lowest quality on QMSum. 

\begin{table}[t]
    \centering
    \small
    \setlength{\tabcolsep}{3pt}
    \begin{tabular}{llllll}
    \toprule
        \textbf{Model} & \textbf{R-1} $\uparrow$ & \textbf{R-2} $\uparrow$ & \textbf{R-L} $\uparrow$ & \textbf{Ent G} $\uparrow$ & \textbf{GPU} $\downarrow$ \\
        \midrule
        Se3 & 38.09 & 11.30 & 36.56 & 0.50 & 11.3 \\
        BlockAttn & 32.01 & \phantom{0}8.99 & 30.90 & 1.61 & 25.7 \\
        Longformer & 42.78 & \textbf{13.21} & 41.34 & 0.97 & 25.3 \\
        LongT5 & 42.03 & 12.67 & 40.76 & 1.03 & 25.4 \\
        Unlimiformer & 35.17 & 11.98 & 34.28 & \underline{1.33} & 27.0 \\
        Extract-Abstract & 19.95 & \phantom{0}5.58 & 19.70 & 0.06 & 13.1 \\
        \midrule
        \model \\
        \quad Attn Only & \hlc[allgreen]{{\textbf{46.05}}}$^\dag$ & \hlc[se3blue]{{\underline{13.09}}}$^\dag$ & \hlc[allgreen]{{\textbf{44.21}}}$^\dag$ & \hlc[se3blue]{{0.81}}$^\dag$ & 13.2 \\
        \quad Attn + Txt & \hlc[allgreen]{{\underline{45.30}}}$^\dag$ & \hlc[se3blue]{{12.63}}$^\dag$ & \hlc[allgreen]{{\underline{43.51}}}$^\dag$ & \hlc[se3blue]{{0.90}}$^\dag$ & 14.2 \\
        \bottomrule
    \end{tabular}
    \caption{
    Results on TV transcripts in SummScreen. We report Ent Graph instead of DiscoScore, as DiscoScore encounters errors when identifying focus. \model with the attentive memory obtains the best R1 and RL scores, while the low accuracy of the extracted salient content leads to performance drop of the summarizer.
    }
    \label{tab:summscreen_result}
\end{table}

This trend is similarly observed on \textbf{SummScreen} (Table~\ref{tab:summscreen_result}) and \textbf{arXiv} (Table~\ref{tab:arxiv_result}), where the extract-then-abstract method performs poorly and adding extracted content leads to performance drop of \model due to the low performance of the extractor.
Meanwhile, \model with the attentive memory is able to obtain the best ROUGE-1 and ROUGE-L scores.
On arXiv, models that use efficient attentions obtain the higher ROUGE scores, because truncating arXiv documents has little effect on summary generation---arXiv articles have the most \textbf{uneven} distributions of salient content, where only about $10\%$ of new salient bigrams are located in the second halves of the documents~\cite{huang-etal-2021-efficient}.

\begin{table}[t]
    \centering
    \small
    \setlength{\tabcolsep}{3pt}
    \begin{tabular}{llllll}
    \toprule
        \textbf{Model} & \textbf{R-1} $\uparrow$ & \textbf{R-2} $\uparrow$ & \textbf{R-L} $\uparrow$ & \textbf{Disco} $\downarrow$ & \textbf{GPU} $\downarrow$ \\
        \midrule
        Se3 & 40.74 & 17.96 & 36.87 & 1.33 & 12.8 \\
        BlockAttn & {\textcolor{boldcolor}{\textbf{49.12}}} & {\textcolor{boldcolor}{\textbf{21.69}}} & {\textcolor{boldcolor}{\textbf{44.40}}} & 1.77 & 25.7  \\
        Longformer & \underline{48.59} & \underline{21.45} & \underline{43.99} & 2.17 & 25.2 \\
        LongT5 & 48.25 & 20.74 & 43.41 & \underline{0.97} & 25.5 \\
        Unlimiformer & 47.78 & 20.58 & 43.22 & 1.22 & 26.8 \\
        Extract-Abstract & 42.37 & 16.43 & 38.62 & 1.03 & 15.3 \\
        PageSum & 46.01 & 18.77 & 41.55 & {\textcolor{boldcolor}{\textbf{0.88}}} & 26.2 \\
        \midrule
        \model \\
        \quad Attn Only & \hlc[se3blue]{42.51}$^\dag$ & \hlc[se3blue]{18.96}$^\dag$ & \hlc[se3blue]{38.56}$^\dag$ & \hlc[se3blue]{1.30} & 16.0 \\
        \quad Attn + Txt & \hlc[se3blue]{44.20}$^\dag$ & \hlc[se3blue]{{18.89}}$^\dag$ & \hlc[se3blue]{40.07}$^\dag$ & \hlc[se3blue]{1.32} & 16.5  \\
        \bottomrule
    \end{tabular}
    \caption{
    Results on arXiv papers. \model variants again outperform Se3. 
    For $80\%$ of the arXiv documents, efficient attention models and PageSum can fully train on their first halves, covering $90\%$ of the salient content that appear in the references~\cite{huang-etal-2021-efficient}, thus the better ROUGE scores than models encoding smaller segments. 
    }
    \label{tab:arxiv_result}
\end{table}


\begin{table}[t]
    \centering
    \small
    \setlength{\tabcolsep}{2pt}
    \begin{tabular}{llllll}
    \toprule
        \textbf{Model} & \textbf{R-1} $\uparrow$ & \textbf{R-2} $\uparrow$ & \textbf{R-L} $\uparrow$ & \textbf{Disco} $\downarrow$ & \textbf{GPU} $\downarrow$ \\
        \midrule
        Se3 & \underline{40.78} & \underline{10.16} & \underline{39.77} & \phantom{0}\underline{10.46} & 11.5 \\
        BlockAttn & 23.45 & \phantom{0}3.09 & 22.09 & 190.27 & 25.7 \\
        Longformer & 20.20 & \phantom{0}2.45 & 18.55 & 204.48 & 25.3 \\
        LongT5 & 33.15 & \phantom{0}6.74 & 32.62 & \phantom{0}24.24 & 25.5 \\
        Unlimiformer & 38.09 & \phantom{0}9.55 & 37.41 & \phantom{0}47.72 & 27.0 \\
        \midrule
        \model (Attn) & \hlc[allgreen]{{\textbf{41.11}}} & \hlc[allgreen]{{\textbf{10.63}}} & \hlc[allgreen]{{\textbf{40.20}}} & \hlc[allgreen]{{\textbf{\phantom{0}10.36}}} & 24.0 \\
        \bottomrule
    \end{tabular}
    \vspace{-2mm}
    \caption{
    Results on novels in BookSum. \model with attentive memory in all layers achieves the best performance on all metrics. Methods requiring external extractors are not included due to the computational cost of building extractive oracles for long novels.
    }
    \label{tab:booksum_result}
    \vspace{-2mm}
\end{table}

Finally, experiments on \textbf{BookSum} show that the divide-and-conquer method produces better summaries for long novels, while our method can further boost its performance (Table~\ref{tab:booksum_result}). However, we find it necessary to incorporate external memory into all layers, suggesting a more complex interaction of external memory with the summarization process for novel plots. Unlike other document types tested, novel plots are typically sequential with less redundancy, which reduces the necessity of the memory mechanism.

%% file: 08_conclusion.tex
\section{Conclusion}
We present \model for summarizing long documents in a memory-constrained setting. 
Based on the divide-and-conquer strategy, \model uses two mechanisms to gather global context and improve summary quality. 
First, external memories on the encoder and decoder are employed to track previously read document content and the corresponding summaries.
Second, the encoder is informed of global salient content predicted by an extractor via text or representation concatenation. 
%
On five summarization datasets, \model generates summaries with better informativeness, faithfulness, and coherence than a baseline divide-and-conquer system. 
Under the same memory constraint, \model outperforms competitive models that leverage efficient attentions or dynamic extraction to preserve global context, highlighting its effectiveness in supplying global context.

%% file: 100_limitations.tex
\newpage
\section{Limitations}


\model's external memory mechanism is restricted to operating solely from past segments to the current segment. This means that the model does not leverage the information contained in future segments, which can be relevant for a comprehensive understanding of the current segment.
To address this limitation, we have designed the global salient content augmentation mechanism to cover context from the future segments, yet more advanced solutions can be explored in future work. For example, on the encoder, making the external memory bidirectional is a potential approach.


While being memory-efficient, the external memory mechanism of \model necessitates a longer running time due to its recurrent nature.
The need for recurrent computations may lead to increased processing requirements, which could impact real-time applications or scenarios where rapid responses are crucial.
The running times of different models are provided in Appendix~\ref{appendix:running_time} for reference. 
Although our model is slower than that of LongT5 and Se3, it still outperforms several other competitive models in terms of speed,
and we will investigate methods for reducing the running time in future work.

%% file: 101_ethics_statement.tex
\section{Ethical Consideration}

We anticipate that one of the major use cases of \model is to allow ordinary users who have computing devices with limited memory to quickly understand government policies and other types of long documents.
However, we recognize that the system generated summaries might not comprehensively cover the salient content that is essential for correctly understanding the policies, causing risks ranging from capital loss to legal liability.
Moreover, system summaries might contain statements that cannot be verified through the document, which further adds to the risks of real-world deployment.
We suggest developers who intend to use our model for real-world application carefully study the outputs by our model before the actual deployment.

%% file: appendix_se3.tex
\section{Divide-and-Conquer Architecture}
\label{appendix:se3}

We choose Se3~\cite{moro2022se3} as our base divide-and-conquer architecture because it can be applied to any document-summary pair. In order to create divide-and-conquer training data for summarization, for each document-summary pair, the document is first divided into segments (\S\ref{subappendix:documet_segmentation}) and each summary sentence is then assigned to a document segment as part of the generation target (\S\ref{subappendix:target_assignment}).

\subsection{Document Segmentation}
\label{subappendix:documet_segmentation}

\begin{algorithm}
\caption{Document Segmentation}
\label{alg:document_segmentation}
\SetAlgoLined

\KwData{Input document $doc$; Segment min, max length $l_{min}$, $l_{max}$}
$segs \leftarrow []$\;
$currSeg \leftarrow []$\;

\ForEach{$sent$ \textbf{in} $doc$}{
    \If{$\text{len}(currSeg) < l_{min}$}{
        $currSeg \leftarrow currSeg + [sent]$\;
    }
    \ElseIf{$\text{len}(currSeg) > l_{max}$}{
        $segs \leftarrow segs + [currSeg]$\;
        $currSeg \leftarrow [sent]$\;
    }
    \Else{
        $nextSeg \leftarrow \text{pseudoSegment}$\;
        \If{$\text{sim}(nextSeg, sent) > \text{sim}(currSeg, sent)$}{
            $segs \leftarrow segs + [currSeg]$\;
            $currSeg \leftarrow [sent]$\;
        }
        \Else{
            $currSeg \leftarrow currSeg + [sent]$
        }
    }
}
$segs \leftarrow segs + [currSeg]$\;
\KwRet{$segs$}

\end{algorithm}

The length of each document segment is between $512$ and $768$ tokens. During segmentation, the algorithm loops through all document sentences, as shown in Algorithm~\ref{alg:document_segmentation}.
A document sentence will be added to the current segment if the segment contains less than $512$ tokens.
The current segment will be finalized if the current segment contains more than $768$ tokens or the current sentence is more semantically similar to the next pseudo segment than the current segment,
where the next pseudo segment is created by including future sentences until reaching $512$ tokens.
To measure the similarity between the current sentence and a segment, we use the average cosine similarity between the representation of the current sentence and representations of the sentences in the segment.
Sentence representations are obtained using Sentence Transformer~\cite{reimers-2019-sentence-bert} with the \texttt{all-roberta-large-v1} model.

\subsection{Target Assignment}
\label{subappendix:target_assignment}

For each sentence in the reference summary, we calculate its ROUGE scores with the document segments. The sentence will then be assigned to the document segment with which yields the highest ROUGE-1 and ROUGE-2 scores.

%% file: appendixA_additional_result.tex
\section{Additional Results}
\label{appendix:additional_result}

\subsection{Running Time}
\label{appendix:running_time}

\begin{figure}
    \centering
    \includegraphics[width=0.48\textwidth]{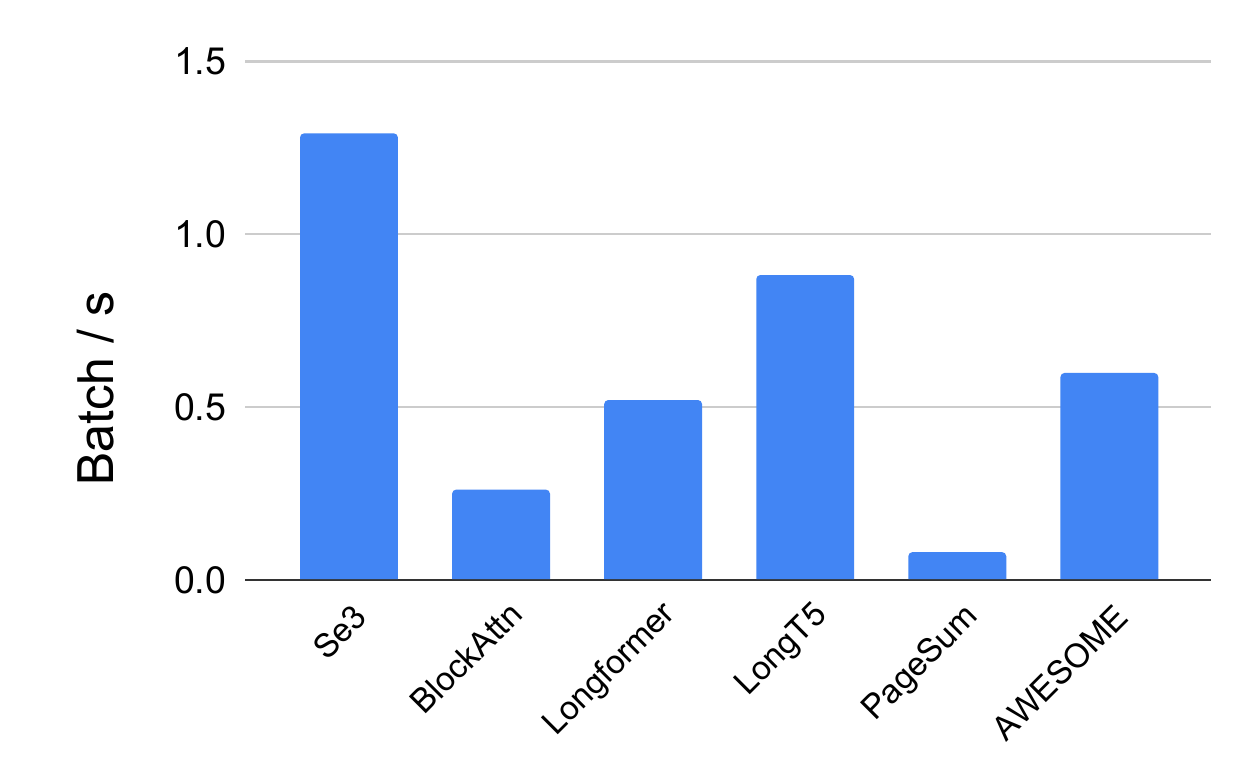}
    \caption{
    Running time (batch per second) of each model. A higher number of batches processed per second indicates a faster running speed. All models use a batch size of 1 and the input is truncated to 16384 tokens.
    }
    \label{fig:running_time}
\end{figure}

We compare the model running time on GovReport (Figure~\ref{fig:running_time}). The input document is truncated to $16384$ tokens and each model is separately train for $1000$ steps with a batch size of 1.
No other computation-heavy program is running at the same time.
While \model take longer time to complete training than Se3, it is still the third fastest model.

%% file: appendixB_dataset.tex
\section{Dataset Details}
\label{appendix:dataset}

\subsection{Statistics}
\label{appendix:dataset_stat}

We conduct experiments on five long document summarization datasets with diverse genres. 
\textbf{GovReport}~\cite{huang-etal-2021-efficient} contains long reports and their summaries written by government research agencies. 
\textbf{QMSum}~\cite{zhong-etal-2021-qmsum} is a query-focused long meeting transcript summarization dataset, with summary-worthy content spread over the documents. 
We prepend the query to all segments. 
We further use a screenplay summarization dataset, \textbf{SummScreen}~\cite{chen-etal-2022-summscreen}, which contains the transcripts of TV series. The TMS subset, with more samples and longer summaries, is selected.
Moreover, we experiment with the scientific papers and their abstracts from \textbf{arXiv}~\cite{cohan-etal-2018-discourse}. 
Finally, we test our models on summarizing \textit{full} novels in \textbf{BookSum}~\cite{kryscinski-etal-2022-booksum}.
For all datasets, we use the official train/dev/test splits if their original data files are released.

Statistics of datasets are reported in Table~\ref{tab:dataset_statistics}.
For GovReport\footnote{\url{https://gov-report-data.github.io/}}, QMSum\footnote{\url{https://github.com/Yale-LILY/QMSum}}, and SummScreen~\cite{chen-etal-2022-summscreen}, we use the data released by the original papers.
For arXiv, we use the version provided by Huggingface Datasets.\footnote{\url{https://huggingface.co/datasets/scientific_papers}}
As the original data files for BookSum are not released due to summary copyright, we use the version reproduced by Unlimiformer~\cite{bertsch2023unlimiformer}.

\begin{table}[t]
    \centering
    \small
    \setlength{\tabcolsep}{4pt}
    \begin{tabular}{lrrrrr}
    \toprule
        & \multicolumn{3}{c}{\textbf{\# Samples}} & \multicolumn{2}{c}{\textbf{\# Word}} \\
        \textbf{Dataset} & \textbf{Train} & \textbf{Dev} & \textbf{Test} & \textbf{Doc} & \textbf{Summ} \\
        \midrule
        GovReport & 17{,}516 & 974 & 973 & 9{,}409 & 553 \\
        QMSum & 1{,}257 & 272 & 279 & 9{,}070 & 70 \\
        SummScreen & 18{,}915 & 1{,}795 & 1{,}793 & 6{,}421 & 381 \\
        arXiv & 203{,}037 & 6{,}436 & 6{,}440 & 6{,}030 & 273 \\
        BookSum & 314 & 45 & 46 & 143{,}301 & 1{,}294 \\
        \bottomrule
    \end{tabular}
    \caption{Statistics of datasets used in our experiments.}
    \label{tab:dataset_statistics}
\end{table}

\subsection{Input Truncation}
\label{appendix:model_input}

\begin{table}[t]
    \centering
    \small
    \setlength{\tabcolsep}{2pt}
    \begin{tabular}{lccccc}
    \toprule
         & \multicolumn{5}{c}{\textbf{Dataset}} \\
        \textbf{Model} & \textbf{Gov} & \textbf{arXiv} & \textbf{QMSum} & \textbf{SumScrn} & \textbf{Book} \\
        \midrule
        Se3 & 50x & 50x & 50x & 50x & 50x \\
        Ext-Abs $^\dag$ & 1x ($\infty$) & 1x ($\infty$) & 1x ($\infty$) & 1x ($\infty$) & - \\
        BlockAttn & 6x & 6x & 8x & 6x & 6x \\
        Longformer & 8x & 8x & 8x & 8x & 8x \\
        LongT5 & 6x & 6x & 6x & 6x & 6x \\
        Unlimiformer & 2x & 2x & 2x & 2x & 2x \\
        PageSum & 3x & 5x & 2x & - & - \\
        \model & 50x & 50x & 50x & 50x & 50x \\
        \bottomrule
    \end{tabular}
    \caption{
    Truncation thresholds (multiply by $1024$) used by each model on different datasets to comply with the memory constraint during training.
    $\dag$: For the extract-then-abstract model, the abstractor has a maximum input length of $1024$, while the extractor can consume all sentences in the document.
    }
    \label{tab:model_input_length}
\end{table}

In our main experiments, we employ a GPU memory constraint of 27GB.
As some baseline models require the input length to be a multiplier of $1024$, setting a constraint of 24GB, a more common number, would lead to further truncation and significant performance drop.

To fit models into our memory constraint, we truncate the model inputs.
The truncation thresholds used by each model on different datasets are shown in Table~\ref{tab:model_input_length}.
Although Se3 and \model theoretically maintain a consistent GPU memory consumption during training regardless of the number of input tokens processed, we have chosen to restrict the maximum number of input tokens in a training sample to $51200$ for reasonable training time.

%% file: appendixC_implementation.tex
\section{Implementation Details}
\label{appendix:implementation}

\paragraph{Baselines.}

BlockAttn and Longformer use block-wise attentions~\cite{phang2022investigating} and sliding-window attentions~\cite{beltagy2020longformer},
where a global token can attend to and be attended by all tokens, while other tokens can only attend to tokens in the same block or window.
LongT5~\cite{guo-etal-2022-longt5} is a sliding-window attention model pre-trained on long sequences,
and Unlimiformer~\cite{bertsch2023unlimiformer} extends BART by selecting input tokens to be attended to via KNN searching.
For the extract-then-abstract approach, we use the same extractor as in the global salient content augmentation of our model, and the abstractor takes as input oracle extracted sentences during training.
Lastly, PageSum~\cite{liu2022page} synthesizes the output representations given by different document segments with dynamic weights.

\paragraph{Extractor.}

The extractor first uses a RoBERTa~\cite{liu2019roberta} to encode each sentence and takes the average of the final layer's outputs as the sentence representation. It then applies a self-attention on top of all sentence representations. The resulting representations are converted to extraction scores after applying a multi-layer perception with one hidden layer.
The extractor is trained with \textbf{oracle extractive labels} that are constructed by greedily searching for document sentences that maximize the sum of ROUGE-1 and ROUGE-2 scores, compared against the reference summary. We do not compute ROUGE-L as in DYLE~\cite{mao-etal-2022-dyle}, because finding the longest common subsequence is computationally expensive and does not yield performance gain.

\paragraph{Training Parameters.}

We train all models with a maximum learning rate of $\num{5e-5}$, except that LongT5 is trained with a maximum learning rate of $\num{1e-4}$. We use a running batch size of $1$ and apply gradient accumulation to achieve an effective batch size of $8$. 
The numbers of training epochs are $3$, $9$, $6$, $2$, $10$ on GovReport, QMSum, SummScreen, arXiv, and BookSum, with warmup steps of $300$, $100$, $300$, $1000$, and $40$.
Due to the computational cost of training long document summarization, each model is trained for a single run.

\paragraph{Model Size.}

\model is based on \texttt{BART-large}\footnote{https://huggingface.co/facebook/bart-large} and has 708 millions of parameters.

\paragraph{Computing Infrastructure.}

All experiments are conducted on RTX A6000 GPUs.

\paragraph{Evaluation Metrics.}

For ROUGE~\cite{lin-2004-rouge}, we use the Python implementation by Google.\footnote{\url{https://pypi.org/project/rouge-score/}}
The official code for DiscoScore~\cite{zhao2022discoscore} is used\footnote{\url{https://github.com/AIPHES/DiscoScore}}, which also provides an implementation of the Ent Graph metric~\cite{guinaudeau-strube-2013-graph}.
We implement the entity precision measure ourselves and run the official code for SummaC~\cite{laban-etal-2022-summac}.\footnote{\url{https://github.com/tingofurro/summac}}
All metrics used are open-source and can be distributed for research purposes.